**Exploring Deep Learning Approaches to Predict Person and Vehicle Trips: An Analysis of NHTS Data**

**Kojo Adu-Gyamfi**
Department of Civil and Environmental Engineering
Institution of Transportation, Ames, Iowa, and 50010
Email: kgyamfi@iastate.edu

**Sharma Anuj**
Professor Civil Construction & Environmental Eng
Institute For Transportation
Ames, Iowa, 50010
Email: anujs@iastate.edu

*Author, Author, and Author*

**ABSTRACT**
Modern transportation planning relies heavily on accurate predictions of person and vehicle trips. However, traditional planning models often fail to account for the intricacies and dynamics of travel behavior, leading to less-than-optimal accuracy in these predictions. This study explores the potential of deep learning techniques to transform the way we approach trip predictions, and ultimately, transportation planning. Utilizing a comprehensive dataset from the National Household Travel Survey (NHTS), we developed and trained a deep learning model for predicting person and vehicle trips.
The proposed model leverages the vast amount of information in the NHTS data, capturing complex, non-linear relationships that were previously overlooked by traditional models. As a result, our deep learning model achieved an impressive accuracy of 98% for person trip prediction and 96% for vehicle trip estimation. This represents a significant improvement over the performances of traditional transportation planning models, thereby demonstrating the power of deep learning in this domain.
The implications of this study extend beyond just more accurate predictions. By enhancing the accuracy and reliability of trip prediction models, planners can formulate more effective, data-driven transportation policies, infrastructure, and services. As such, our research underscores the need for the transportation planning field to embrace advanced techniques like deep learning. The detailed methodology, along with a thorough discussion of the results and their implications, are presented in the subsequent sections of this paper.

**INTRODUCTION**

Transportation planning is a multifaceted and complex field with far-reaching influence on numerous societal aspects, from economic development to environmental sustainability (*1*). At the heart of this planning process lies the accurate prediction of person and vehicle trips, providing the groundwork for infrastructure development, policy making, and service planning (*2*).

Historically, traditional transportation planning models have been relied upon to predict these trips. However, these models often operate under certain assumptions and typically rely on simplistic statistical methods, leading to an under-representation of the complexity and dynamic nature of travel behavior (*3*). The resulting inaccuracies in trip predictions can culminate in ineffective planning and misguided resource allocation (*4*).

In this era of big data, the surge in available travel data, such as that provided by the National Household Travel Survey (NHTS), introduces opportunities for more nuanced understanding of travel behavior (*5*). Concurrently, the advent of more sophisticated computational techniques, particularly deep learning, offers the tools to unlock this data's potential. Deep learning models have the capacity to discern complex, non-linear patterns from large-scale datasets, making them ideal candidates for modeling the intricate dynamics of travel behavior (*6*).

This paper explores the application of deep learning for predicting person and vehicle trips using NHTS data. We aim to demonstrate that deep learning techniques, renowned for their ability to model complex relationships and patterns in data, can produce more accurate trip predictions than traditional planning models. The results of this study hold significant implications for transportation planning, laying the foundation for more effective, data-driven decision-making (*7*).

We present a deep learning model developed for this purpose, detailing its design, training process, and evaluation. Our model notably achieved an accuracy of 98% for person trip prediction and 96% for vehicle trip estimation, significantly surpassing the performance of traditional models.

The paper is organized as follows: Section 2 provides a review of relevant literature on traditional transportation planning models and deep learning applications in transportation. Section 3 outlines the methodology, encompassing data description, deep learning model design, and the training process. Section 4 presents the results and the performance evaluation of our model, and discusses the implications of our findings for transportation planning, and finally, Section 5 concludes the paper, offering recommendations for future research.

**LITERATURE REVIEW**
Transportation planning models have a rich history and have been widely used in predicting travel demand, facilitating the development of infrastructure, and informing policy decisions (*8*). These models primarily adopt statistical approaches, with the Four-Step Model (Trip Generation, Trip Distribution, Mode Choice, and Route Assignment) being one of the most iconic (*18*).

Despite their usefulness, the inherent assumptions within these models have often been critiqued. They usually consider travel decisions as independent entities, whereas in real scenarios, these decisions are interconnected. The Four-Step Model, for instance, often fails to capture the cascading effect of one decision onto the next (*10*). This model, as well as others, tends to be static, overlooking temporal changes and individual heterogeneity in travel behavior (*11*). Such simplifications could lead to misrepresentation and inaccuracies in trip predictions, thereby misguiding resource allocation and transportation planning.

In recent years, the proliferation of big data in transportation has brought about opportunities for a more refined understanding and modeling of travel behavior (*12*). Comprehensive datasets, like the National Household Travel Survey (NHTS), capture a wide array of details about individual and household travel behavior (*13*). These data assets offer the potential for more complex and accurate modeling of travel behavior, thus opening avenues for novel approaches, including deep learning.

Deep learning, a powerful subset of machine learning, is particularly adept at unraveling intricate patterns in vast and high-dimensional datasets (*14*). It has seen increasing adoption in the transportation domain, with tasks like traffic flow prediction (*15*), mode choice modeling (*16*), and route choice modeling (*17*) being tackled effectively. These applications underscore the prowess of deep learning in learning from and predicting complex patterns, thereby enhancing accuracy.

The past decade has also seen an upsurge in transportation research exploring the realm of travel demand modeling using artificial intelligence techniques. Researchers have begun to explore how machine learning can contribute to a more refined understanding of factors influencing travel demand and predicting future trends (*19*). Despite these advances, the potential of deep learning in person and vehicle trip prediction in transportation planning remains largely untapped.

As such, this paper endeavors to fill this research gap by proposing a deep learning model for predicting person and vehicle trips, utilizing the wealth of information present in the NHTS data. We believe this approach will prove instrumental in improving the accuracy of predictions, thereby contributing to more efficient and effective transportation planning.

**METHODS**
The proposed methodology for this research focuses on a comprehensive data exploration of the National Household Transportation Survey (NHTS) data, followed by feature selection, model development, training, and testing.

**Data**
The foundational base of this research lies in the NHTS 2017 dataset (*20*), which serves as the primary source for training, validation, and testing. This data is extensively categorized into diverse segments, such as household details, demographic characteristics, vehicle ownership, trips taken, and mode choice, thereby offering an inclusive scope for model training (*21*). The dataset spans across 129,696 households, classified under six geographic groups, further subdivided into urban, suburban, and rural areas. **Figure 1** provides a detailed structure of the NHTS data.

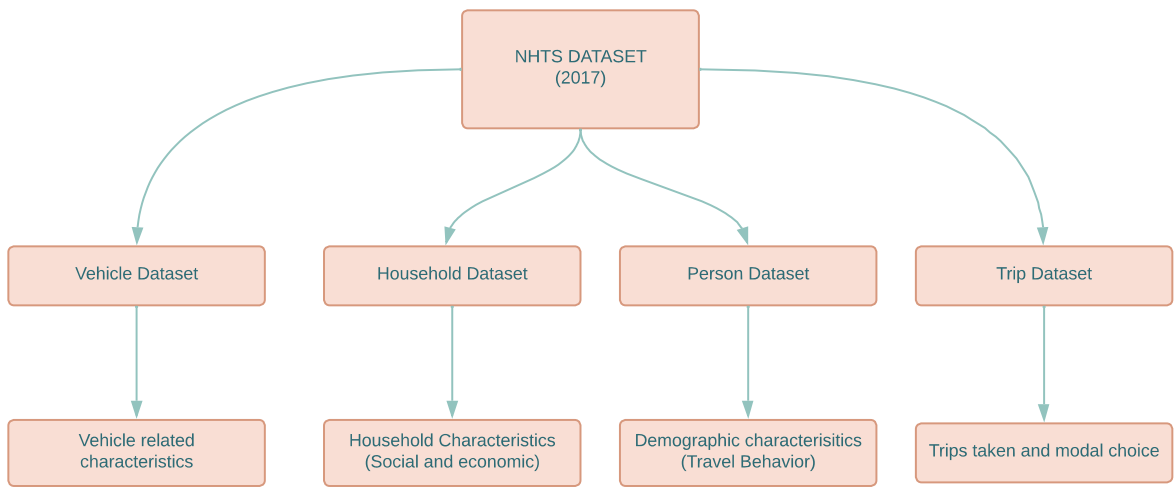

**Figure 1 NHTS data structure**

**Response Variable and Independent Variables**
The research aims to leverage neural network machine learning methodologies to forecast person and vehicle trips as defined by the NHTS (*10*). The person and vehicle trips serve as the response variables for the study. The independent variables, listed in **TABLE 1**, are chosen based on previous research emphasizing their potential influence on trip generation (*22,23*).

**TABLE 1 Independent Variables**

| Variable | Description |
|---|---|
| 1. Household Vehicles (0 available) | Households with zero vehicles count |
| 2. Household Vehicles (1 available) | Households with one vehicle count |
| 3. Household Vehicles (2+ available) | Households with two or more vehicles counts |
| 4. Workers in Household (1 available) | Households with only one worker |
| 5. Workers in household (2+ available) | Households with two or more workers |
| 6. Total Population | Total population of the community based on the five-year American Community Survey |
| 7. Life Cycle (1+ child <18) (Count) | Count of one or more children in household, less than 18 years old. |
| 8. Life Cycle (1 person household, <65) (Count) | One person household, less than 65 years old |
| 9. Life Cycle (2+ person household, 0 65+) (Count) | Two or more-person household all less than 65 years old |
| 10. Life Cycle (1 person household, 65+) (Count) | One person household, greater than 65 years old |
| 11. Life Cycle (2+ person household, 1+ 65+) (Count) | Two or more-person household with aleast one 65 or more years of age |
| 12. Population in group quarters | Total population in groups |
| 13. Count of households | Total count of households' members |
| 14. Household Income | Household income categories |
| 15. Cluster | Census Region/ Division(geographic areas) |
| 16. Urban Group | Level of urbanicity |

**Machine Learning Method - Neural Networks**
This research employs neural networks, a subfield of deep learning, due to their proven capability in handling complex problems (*24*). Mimicking the human brain's functioning, these networks progressively learn and improve their performance with minimal human intervention. Neural networks are generally composed of three interconnected layers - the input layer, the hidden layer(s), and the output layer, as depicted in **Figure 2** (*6*).

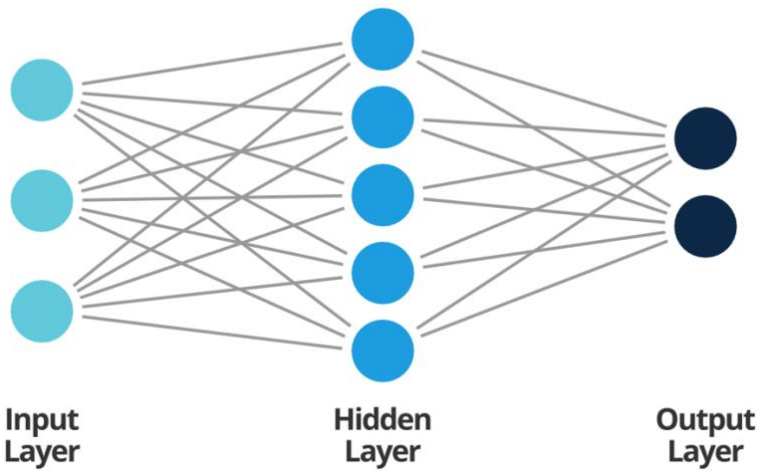

**Figure 2 Neural Network Architecture**

**Deep Learning Implementation**
The research adopts a deep learning neural network approach for trip prediction. Contrasting with basic neural networks, deep learning networks incorporate several hidden layers with millions of neurons, enabling mapping of diverse inputs to diverse outputs. However, these networks require considerable data and longer training durations (*25*).
The deep learning implementation process for trip prediction comprises several steps:

**1. Data Processing**
Data processing commences with eliminating null or empty attributes from the dataset. In this study, data entries devoid of values under the following features were discarded: Urban Group, Estimated person trips, Estimated vehicle trips, Household count, and Median household income. Approximately 3% of the data was thus removed.

The data was subsequently scaled to a standard scale, which ensures faster convergence of the optimization technique. **Figure 3** provides a graphical representation of data counts post-processing.

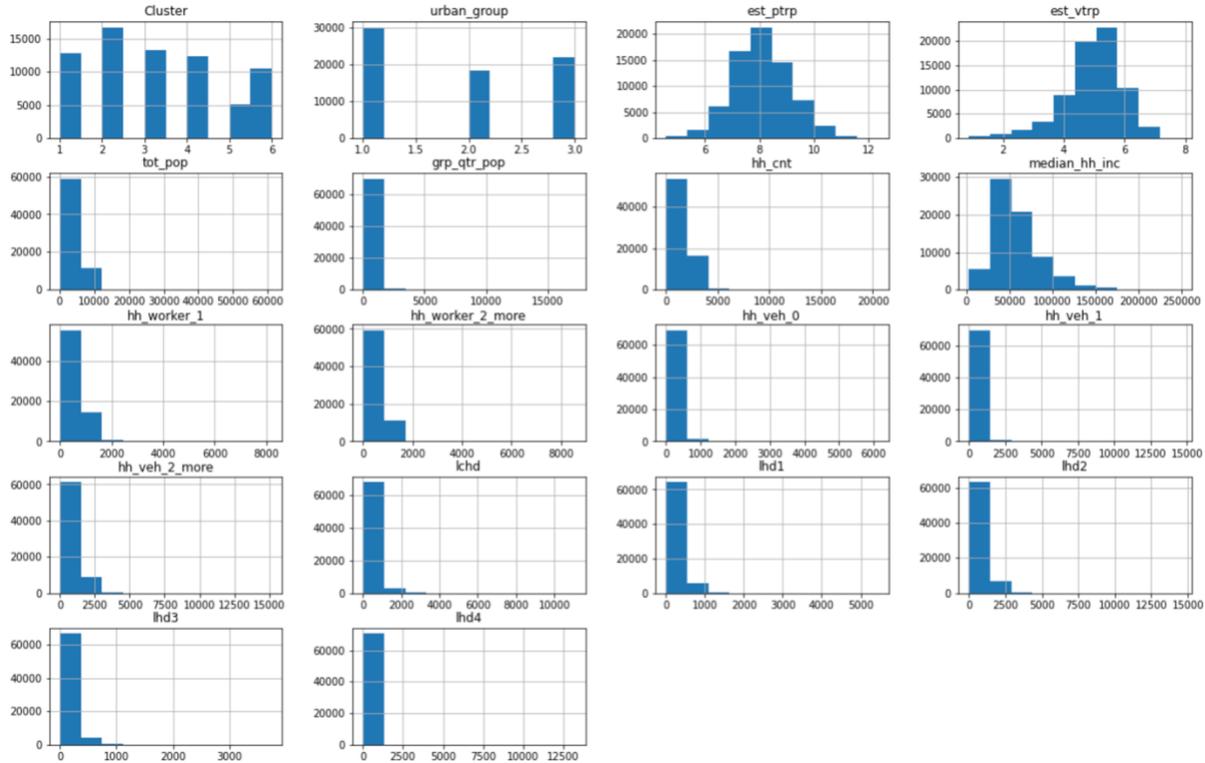

**Figure 3 Data Counts description**

## 2. Designing the Deep Neural Network

The deep learning network design involved determining the total number of neurons and layers. The model employs one input vector per feature, with the Rectified Linear Activation (ReLU) function used on the input layer. It also includes a hidden layer utilizing the tanh activation function, with five nodes, and an output layer containing a single fully connected node to provide the output for the regression task. The final model summary is given in **Figure 4**.

```
Layer (type)                 Output Shape              Param #
=================================================================
dense_3 (Dense)              (None, 5)                 85

dense_4 (Dense)              (None, 5)                 30

dense_5 (Dense)              (None, 1)                 6

=================================================================
Total params: 121
Trainable params: 121
Non-trainable params: 0
```

**Figure 4 Model Summary**

## 3. Training the Deep Neural Network

The training phase involved hyperparameter optimization, where parameters such as batch size and the number of epochs were tuned using GridSearch cross-validation from the Scikit-learn Python library (*26*). **Figure 5** shows the best parameters obtained for the model, the best parameters were a batch size of 20 and epochs of 5. During training, the model utilizes forward and backward propagation to learn and adjust weights of hidden layers.

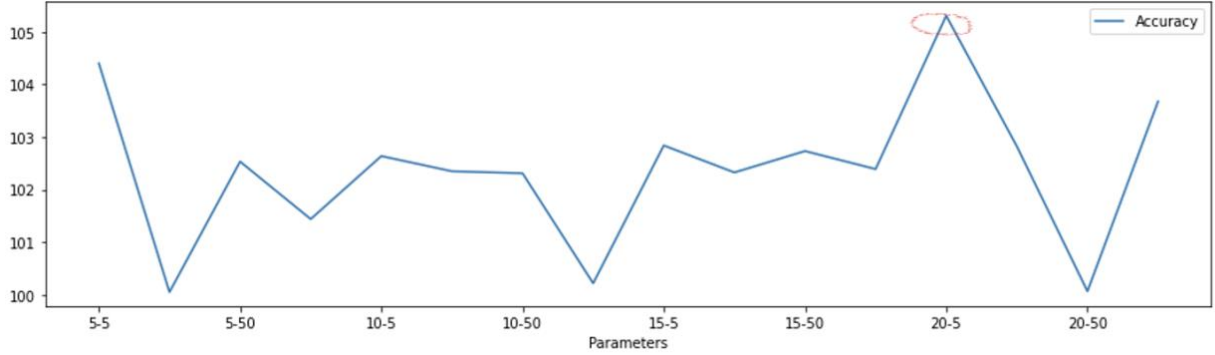

**Figure 5 Hyperparameters (red circle indicates the point for selction of hyper parameters)**

## 4. Testing the Deep Neural Network

The model was evaluated using the Mean Absolute Percentage Error (MAPE) metric (*27*). A subset of the dataset was used for this testing phase. It is expressed as a ratio defined in **Equation 1**

$$\text{MAPE} = \frac{100\%}{n} \sum_{t=1}^{n} \left| \frac{A_t - F_t}{A_t} \right| \qquad (1)$$

**RESULTS AND DISCUSSION**
The results and discussion section delineates the output obtained from the deep learning model trained to predict person and vehicle trips, derived from the NHTS data. The section is in two subsections: Person Trips and Vehicle Trips, as shown below:

**Person Trips**
The outcome of the person trip prediction model is promising, as is evident from the training and validation loss curves depicted in **Figure 6**. A model is considered to be a good fit when the training and validation losses converge to a point of stability with minimal disparity, termed the generalization gap. In this study, the validation and training loss curves exhibit such a trend, thereby affirming the effective learning and generalization capability of the model.

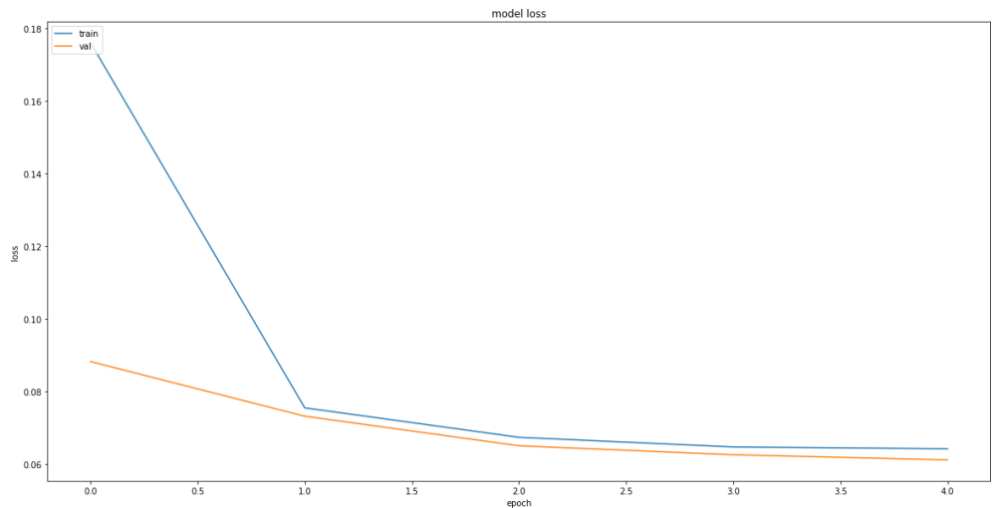

**Figure 6 Person trips loss curves**

The performance of the model, quantified using the Mean Absolute Percentage Error (MAPE), attests to its reliability, with an achieved accuracy rate of 98%. In other words, the model prediction varied by only 2% from the actual values on average. **Figure 7** presents a comparative representation of the actual and predicted person trips, substantiating the model's impressive performance. A close observation of **Figure 8** reveals that the model's predicted values closely follow the actual values, with minimal deviation, further reinforcing the model's effectiveness.

*Author, Author, and Author*

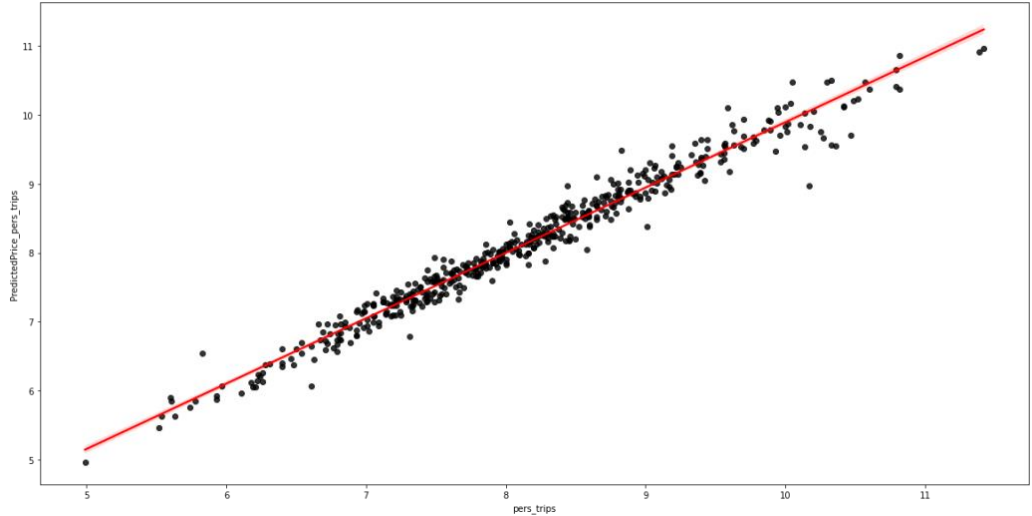
**Figure 7 Person trips Actual vs. predicted regression line**

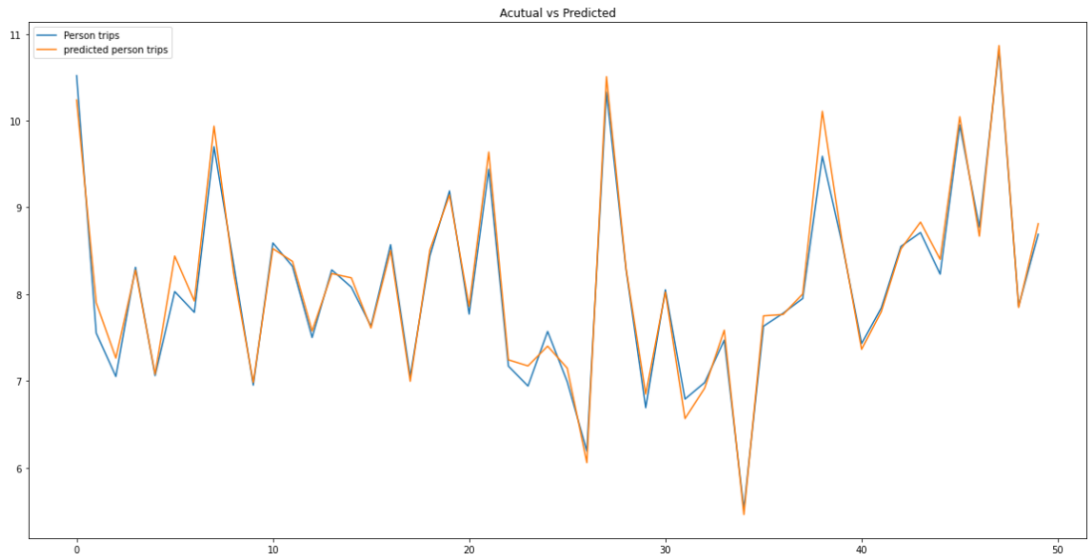
**Figure 8 Person trips Actual (blue colored) vs. predicted curves (Orange Colored).**

**Vehicle Trips**
The vehicle trip prediction model exhibits a similar pattern of convergence of the training and validation loss curves, as shown in **Figure 9**. There are, however, occasional instances where the validation loss slightly exceeded the training loss, hinting at potential overfitting. Despite these occasional divergences, the overall trend suggests that the model learned effectively and is capable of generalization.
The vehicle trip prediction model exhibited an accuracy rate of 96%, as assessed by the MAPE metric, implying a 4% average deviation from the actual values. This level of accuracy further strengthens the model's credibility. As demonstrated in **Figure 10**, the distribution of the model's predicted values for vehicle trips closely follows the distribution of the actual values. A deeper examination of **Figure 11** showcases the minimal margin of error between the actual and predicted values, thus testifying to the model's commendable performance.

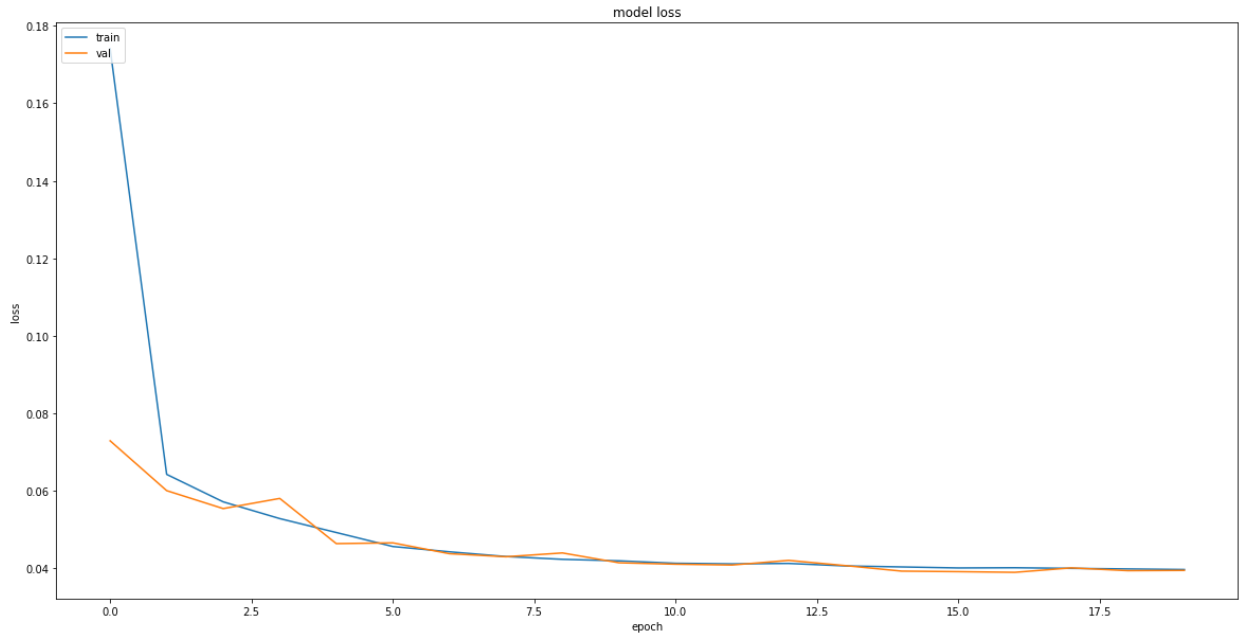

**Figure 9 Vehicle trips lose curves.**

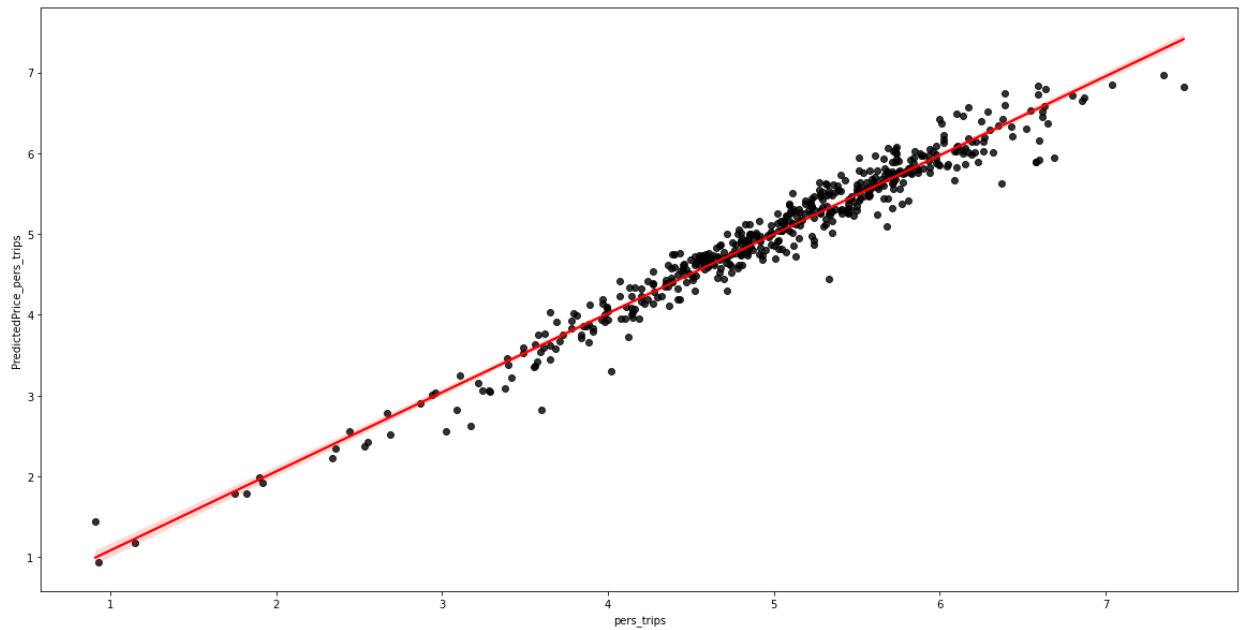

**Figure 10 Vehicle trips Actual vs. predicted regression line.**

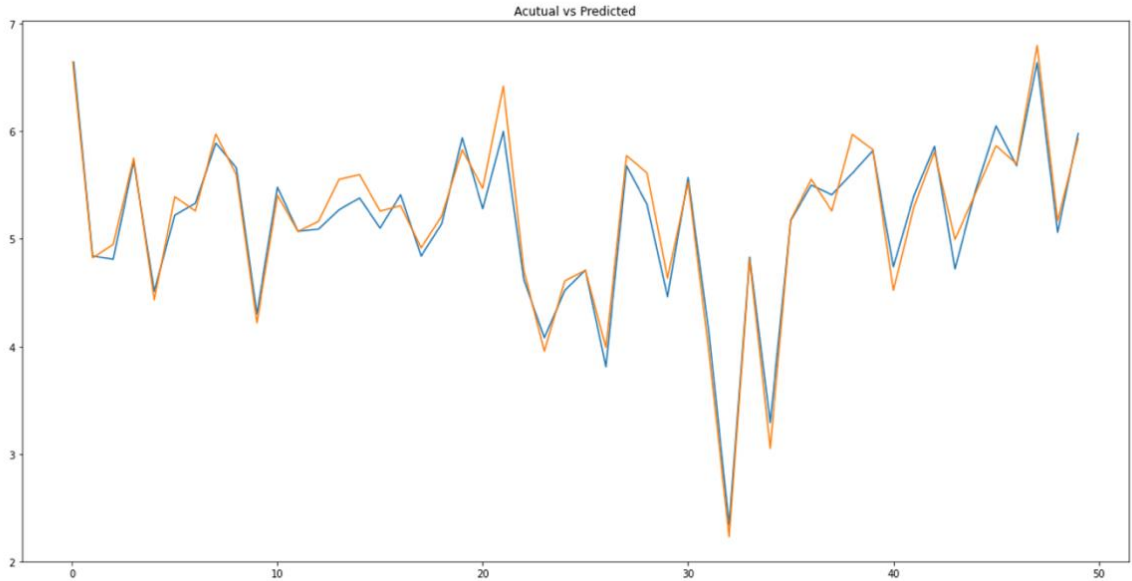

**Figure 11 Vehicle trips Actual (blue colored) vs. predicted curves (orange colored).**

In conclusion, the person and vehicle trip prediction models exhibit commendable performance, delivering high accuracy rates. Despite occasional indications of overfitting in the vehicle trip model, both models effectively learned from the training data and demonstrated robust generalization capabilities, promising their practical utility in trip prediction tasks.

*Author, Author, and Author***CONCLUSIONS**
This study presents an insightful exploration into the domain of predicting person and vehicle trips using deep learning models trained on the National Household Travel Survey (NHTS) data. It demonstrates how modern artificial intelligence tools can significantly contribute to improving our understanding of travel behavior, enabling us to forecast future transportation needs with an enhanced level of accuracy. Our deep learning models for person trip and vehicle trip prediction both exhibited high performance, with accuracy levels of 98% and 96% respectively. These findings, although slightly marred by indications of overfitting in the vehicle trip model, provide substantial evidence of the efficacy of machine learning algorithms in this field. Notably, the models displayed a minimal deviation from actual values, underscoring their robustness and reliability in accurately predicting person and vehicle trips. Furthermore, the methodology applied in this research, characterized by the effective use of training and validation loss curves for performance evaluation and error minimization, offers a valuable blueprint for future investigations in this area. The results are promising, but there is room for enhancement, particularly in addressing the occasional instances of overfitting and refining the prediction accuracy.
In the broader perspective, the successes of this study pave the way for more extensive use of artificial intelligence in the realm of transportation and mobility studies. The predictive models developed here can provide crucial inputs for policy formulation and strategic planning, with implications for environmental sustainability, urban design, and traffic management. Moreover, they hold the potential to improve transport models by integrating them into traditional four-step models or activity-based models, thus enhancing our ability to predict and respond to the shifting paradigms of travel behavior.
In conclusion, while the journey of artificial intelligence in transportation studies is still in its early stages, this research offers a compelling demonstration of the potential it holds. With further refinement and expanded applications, such deep learning models could revolutionize how we predict and plan for future transportation needs.